\documentclass[journal]{IEEEtran}
\usepackage{graphicx}
\usepackage{subfigure}
\usepackage{caption}
\usepackage{hyperref} 
\usepackage{microtype}
\usepackage{makecell}
\usepackage{amsmath}
\usepackage{amsfonts}
\usepackage{bm}
\usepackage{color}
\usepackage{multirow}
\usepackage{etoolbox}
\usepackage{cite}
\usepackage{threeparttable}
\usepackage{booktabs}
\usepackage{flushend}
\usepackage{mathtools}
\usepackage{algorithmicx}
\usepackage{algorithm}
\usepackage[noend]{algpseudocode}

\begin{document}


\title{HILONet: Hierarchical Imitation Learning from Non-Aligned Observations}


\author{Shanqi~Liu, Junjie~Cao, Wenzhou~Chen, Licheng~Wen, and Yong~Liu
	\thanks {Shanqi~Liu, Junjie~Cao, Wenzhou~Chen, Licheng~Wen, and Yong~Liu are with the State Key Laboratory of Industrial Control Technology and Institute of Cyber-Systems and Control, Zhejiang University, Zhejiang, 310027, China, {\tt\small yongliu@iipc.zju.edu.cn}). 

	\textbf{This work has been submitted to the IEEE for possible publication. Copyright may be transferred without notice, after which this version may no longer be accessible.}}

} 


\markboth{Journal of IEEE Transactions on System, Man, and Cybernetics: Systems}%
{Shell \MakeLowercase{\textit{et al.}}: Bare Demo of IEEEtran.cls for IEEE Journals}

\maketitle
\begin{abstract}
	It is challenging learning from demonstrated observation-only trajectories in a non-time-aligned environment because most imitation learning methods aim to imitate experts by following the demonstration step-by-step. However, aligned demonstrations are seldom obtainable in real-world scenarios. In this work, we propose a new imitation learning approach called Hierarchical Imitation Learning from Observation(HILONet), which adopts a hierarchical structure to choose feasible sub-goals from demonstrated observations dynamically. 
	Furthermore, our method can learn different kinds of policy corresponding to different tasks by using a novel reward structure.
	We also present three different ways to increase sample efficiency in the hierarchical structure. We conduct extensive experiments using several environments. The results show the improvement in both performance and learning efficiency.
\end{abstract}


\IEEEpeerreviewmaketitle

\section{Introduction}
Robots can acquire complex behavior skills suitable for various unstructured environments through learning. Two of the most prevalent paradigms for behavior learning in robots are imitation learning(IL) and reinforcement learning(RL). RL methods can theoretically learn behaviors that are optimal with respect to a clear task reward. However, it usually takes millions of training steps to converge. IL methods, on the other hand, can learn faster by mimicking expert demonstrations. But in many real world scenarios, the demonstrations are hard to obtain as we may not be able to obtain expert's actions or the expert has a different action space. In such a case, the more-specific problem of imitation learning from observation(ILfO) must be considered. ILfO refers to only using the observation of demonstrated trajectories to imitate expert, and does not require expert actions label or external rewards.


Previous works in ILfO focus on mimicking an expert skill by following the demonstration step-by-step, such as OptionGAN\cite{Henderson2018} and TCN\cite{Sermanet2018}. They expect the agent can achieve the same observation of the demonstrated trajectory at every time step.
Such methods require assuming that the demonstration of the learning task can be temporally aligned with the agent's actions.
However, this assumption does not hold when environments do not have a constant number of steps to end, which is common in real world scenarios, especially when demonstrator and learning agent have different physical abilities or in different environments. In such a case, it is infeasible to follow the demonstrated trajectory at every time step.
Moreover, these methods can only utilize one demonstrated trajectory, as multiple trajectories will provide multiple states at each time step and the agent can not choose which one should be followed. This leads to lower utilization efficiency of expert trajectories.
As a result, there are a few works focusing on making the non-time-aligned demonstration be time aligned\cite{kim2019cross}, \cite{liu2019state}. However, instead of strictly following a demonstration step-by-step, a more natural way is to select the observations that are feasible to achieve adaptively.
In this case, the key to ILfO is how to choose the feasible goals. This task is similar to the goal-based hierarchical reinforcement learning's(HRL) task.
However, although few works\cite{Lee2019} are drawing on hierarchical reinforcement learning and use it in imitation learning, they did not use it to choose the sub-goal from all demonstrated trajectories.

In our work, we propose a novel hierarchical reinforcement learning method that can flexibly choose an observation from the expert's trajectories and use it as the sub-goal to follow. The whole structure of our method consists of two-part, high level policy and low level policy. High level policy outputs the sub-goal chosen from expert trajectory every a few steps and low level policy takes it as the sub-goal to achieve.
The insight is that when human imitate from observation, we tend to choose some key position to achieve and how we get there is not essential. This means the agent does not have to focus on every step of the expert. Instead, it searches for the observations in demonstrations that can actually be achieved from the current observation. Then the agent only needs to interact with the environment and find the path to achieve the goal.

\begin{figure*}[t]
	\centering
	\includegraphics[width=1.0\textwidth]{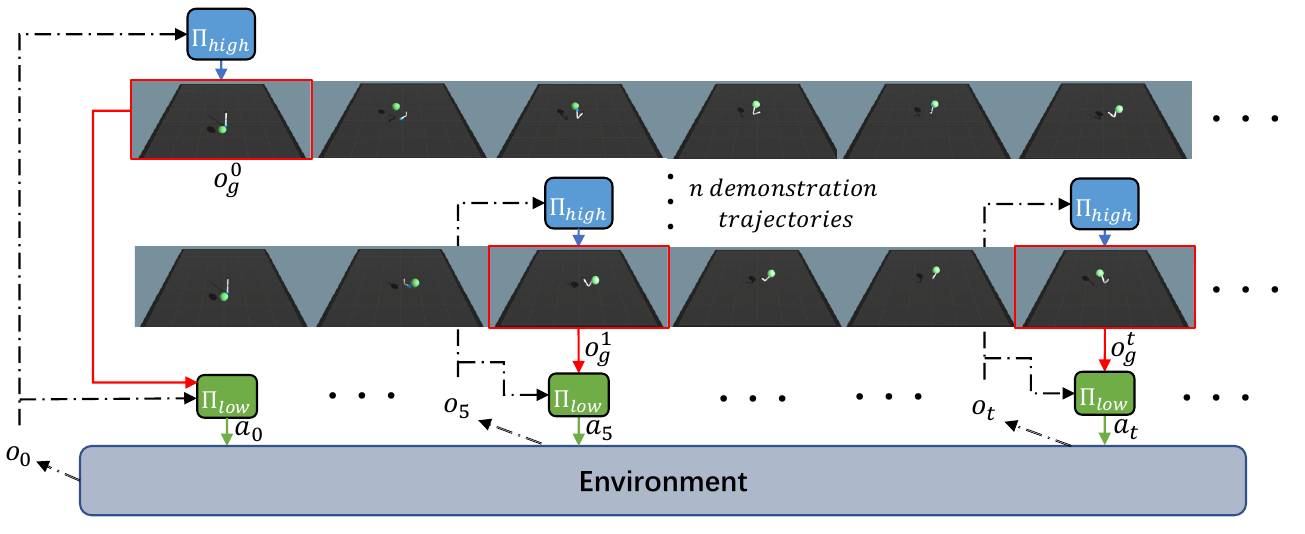}
	\caption{The structure of our overall policy. The high level policy is charged for choosing a reachable sub-goal depending on current observation from all demonstrated trajectories. Low level policy is capable of achieving that sub-goal in specific steps. The rollout is shown in Algorithm. \ref{algo}.}
	\label{structure}
\end{figure*}

Furthermore, to the best of our knowledge, most tasks in real world scenario can be classified into two categories. One can be effectively described by a single goal observation such as navigation. As well as the other kind of tasks such as swimming, which are usually described by a sequence of key observations rather than a single goal. As agent must go through a series of specific position to gain high speed when swimming.
It is clear that we want the agent explores more novel paths to the goal position in the tasks described by a single goal observation and mimics the expert behavior as closely as possible in the tasks described by a sequence of key observations. In our method, we propose a novel reward structure that can control whether the agent follows the trajectories as closely as possible or explores more.
We theoretically prove that our method can solve all two kinds of tasks by control the behavior mode of the agent. Thus, our method has broad applicability.




Moreover, to increase the sample efficiency, we propose several methods to overcome the non-stationarity in hierarchical structure. Firstly, we use a hindsight replacement method to transfer the non-optimal transitions of high level policy in HRL into optimal ones. We also propose a time-delay training method to stabilize the low level policy. Additionally, we choose differentiated experience pools for high level policy and low level policy.

Finally, we test our method and the state-of-art imitation learning methods in five different environments including both single-goal tasks and sequenced-goal tasks. The result indicates our method outperforms all compared methods.
And we demonstrate that our method can solve all two kinds of tasks by control the behavior mode of the agent.

In summary, the main contributions of our work include:
\begin{itemize}
	\item[$\bullet$] We propose a new way of learning from observation using hierarchical reinforcement learning structure to choose feasible sub-goal, which can solve all kinds of non-time-aligned environments.
	\item[$\bullet$] We increase the sample efficiency of hierarchical reinforcement learning by overcoming the non-stationarity.
	\item[$\bullet$] We test our method and SOTA imitation learning method in five different tasks. And we show the ability of our method to solve different kinds of tasks.
\end{itemize}

\section{Related Work}
\subsection{Imitation Learning from Observation(ILfO)}

There are two general groups of ILfO: model-based algorithms, and model-free algorithms\cite{Torabi2019}. Model-based approaches to ILfO are characterized by the fact that they learn some dynamics model during the imitation process. As for model-based part, \cite{Torabi2018} have proposed an algorithm, behavioural cloning from observation(BCO), that is learned an inverse dynamics model using an exploratory policy, and then uses that model to infer the actions from the demonstrations and to use behaviour clone\cite{bain1995framework} to learn. There is also another ILfO approach that learns and uses forward dynamics like imitating latent policies from observation (ILPO) \cite{edwards2019imitating}.

In this work, we mostly discuss model-free algorithms. Adversarial approaches to ILfO are inspired by the generative adversarial imitation learning (GAIL)\cite{ho2016generative} algorithm, but takes ${(o_t, o_{t+1})}$ instead of ${(o_t, a_t)}$ in GAIL. This is called GAILfO\cite{torabi2018generative}. There are other adversarial approaches like \cite{merel2017learning}, \cite{stadie2017third}.

Another type of model-free algorithm is reward engineering.
It means that, based on the expert demonstrations, a manually-designed reward function is used to find imitation policies via RL.
Reward engineering methods use Euler distance of the same time step's policy observation and expert observation to design rewards, like in TCN and \cite{Liu2018}.
Another approach of this type is \cite{goo2019one} in which the algorithm uses a formulation similar to shuffle-and-learn \cite{misra2016shuffle} to train a classifier that learns the order of frames in the demonstration.
\subsection{Hierarchical Reinforcement Learning(HRL)}
There are several RL approaches to learning hierarchical policies\cite{sutton1999between}; \cite{kulkarni2016hierarchical}; \cite{bakker2004hierarchical}. However, these have many strict limits and are not off-policy training methods. Recently popular works like HIRO\cite{nachum2018data}, Option-Critic\cite{bacon2017option} and FeUdal Networks\cite{vezhnevets2017feudal}have achieve quite good performance. Especially, HAC\cite{levy2017learning} using hindsight\cite{andrychowicz2017hindsight} to overcome non-stationarity in HRL and makes off-policy method have a better performance. In this work, we also use hindsight in a different way to overcome non-stationarity in the form of imitation learning. As for using HRL in imitation learning, there are few works, too. Like \cite{Lee2019} using HRL to choose which observation of the expert trajectory should be jumped. However, it still follows a single trajectory step-by-step essentially.

In the hierarchical framework, we consider agent with a natural two-level hierarchy. The high level corresponds to choosing sub-goals, and the low level corresponds to achieve those sub-goals. This structure is typical in past works\cite{konidaris2009skill}.

\section{Background}
\subsection{Markov Decision Process (MDP)}
In standard Markov Decision Process (MDP), one agent sequentially chooses an action $ a_{t} $ according to a policy $ \pi(a|o) $ based on the observation $ o_{t} $ at time t. After taking the action $ a_{t} $, observation $ o_{t} $ transforms to the next observation $ o_{t+1} $ according to the transition probability which satisfies Markov property and is entirely determined by the observation-action pair one time step before, i.e. $ o_{t+1}\sim p(o_{t+1}|o_{t},a_{t}) $. Then the agent receives a scalar reward signal $ r(o_{t},a_{t}) $ from the environment. Deep Reinforcement Learning is one kind of deep learning algorithms that finds a policy $ \pi $ which can extract features with deep neural network and maximizes the expected discounted cumulative reward in one episode, i.e. $ R(o_{t})=E\left[ \sum_{t}\gamma^tr(o_{t},a_{t})\right] $, where $ \gamma $ is a discount factor.
\subsection{Hindsight Experience Replay(HER)}
As we use the off-policy learning algorithm, we also use Hindsight Experience Replay(HER)\cite{andrychowicz2017hindsight}. HER is a data augmentation technique that can accelerate learning in sparse reward tasks. HER first creates copies of the {observation, action, reward, next observation, goal} transitions that are created in traditional off-policy RL. In the copied transitions, the original goal element is replaced with a observation that was achieved during the episode, which guarantees that at least one of the HER transitions will contain the sparse reward. This method can help us to overcome non-stationarity in HRL.
\subsection{Deep Deterministic Policy Gradient(DDPG)}
Policy gradient methods maximize the expected cumulative reward by estimating the performance gradient with respect to the policy parameter vector $ \theta $: $ \nabla_{\theta}J(\pi_{\theta}) $, and updating the policy parameter vector with gradient ascent.

As the original version of policy gradient algorithm, REINFORCE \cite{reinforce} tends to be of high variance due to the gradient estimation with Monte Carlo method:
\begin{equation}
	\label{eq_ac0}
	\nabla_{\theta}J(\theta)\approx\frac{1}{m}\sum_{i=1}^{m}\sum_{t=0}^{N-1}\nabla_{\theta}\log \pi_\theta(a_t|o_t)R_t,
\end{equation}
where $ R_t $ represents the cumulative reward from time t to the end of one episode. Actor-critic methods use the value function $ V(o_t) $, action-value function $ Q(o_t,a_t) $ or advantage function $ A(o_t,a_t)=Q(o_t,a_t)-V(o_t) $ to substitute for the cumulative reward $ R_t $, so as to reduce the variance of gradient estimation and improve the performance of policy gradient methods.
Policy gradient methods maximize the expected cumulative reward by estimating the performance gradient with respect to the policy parameter vector $ \theta $: $ \nabla_{\theta}J(\pi_{\theta}) $, and updating the policy parameter vector with gradient ascent.

For deterministic policy such as DPG \cite{dpg} and DDPG \cite{lillicrap2015continuous}, according to the deterministic policy gradient theorem \cite{dpg}, the gradient of the objective $ J(\theta) $ can be estimated as:
\begin{equation}
	\label{eq_ddpg0}
	\nabla_{\theta}J(\theta)\approx\frac{1}{m}\sum_{i=1}^{m}\sum_{t=0}^{N-1}\nabla_{\theta}\mu_\theta(a_t|o_t)\nabla_{a}Q^{\mu}(o_t,a)|_{a=\mu_\theta(o_t)}.
\end{equation}

\subsection{Notations}
In order to show the algorithm more clearly, we list all the main notations used in our method below.
We first define $D=\{\tau_1,\tau_2,...,\tau_N\}$ means all $N$ trajectories we used in training process.
$\tau_i=\{d_1^i,d_2^i,...,d_{T_i}^i\}$ is one trajectory of demonstrations. $d_j^i$ means the observation of time step $j$ in the number $i$ trajectory of demonstrations. $T_i$ means the length of trajectory $i$.
Then, we define $I(d_j^i)$ to represent the index of $d_j^i$ in its own trajectory.
\begin{equation}
	I(d_j^i) = \frac{j}{T_i}
\end{equation}

\begin{figure*}
	\centering
	\includegraphics[width=1.0\textwidth]{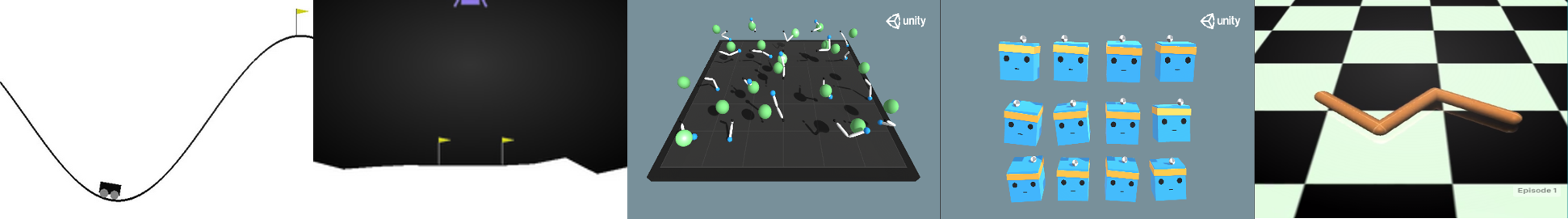}
	\centering
	\caption{All environments used in experiments. From left to right: MountainCar; LunarLander; Reacher:a robot arm try to reach the goal position; 3Dball:a platform robot try to balance the ball; Swimmer.}
	\label{fig_env}
\end{figure*}

\section{Method}
In this paper, we focus on solving imitation learning from observation in non-time-aligned settings. We propose a hierarchical RL framework that learns to find a feasible demonstration observation that the agent can achieve from its correct state. Following this guide, the agent can solve the target task without knowing the exact environment reward. Furthermore, we propose a novel reward structure which can control the behavior mode of the agent. Then, we propose several methods to overcome the non-stationarity in the hierarchical framework to increase sample efficiency.

\subsection{Hierarchical Imitation Learning From Observation Framework}

In our approach, the whole policy is consisted of two part, high level policy $\pi_{high}(o_g|o_t;\theta_h)$ and low level policy $\pi_{low}(a|o_t,o_g;\theta_l)$, where $o_g$ is sub-goal chosen from expert trajectory observations for low level policy, $o_t$ is current observation. We use DDPG algorithm to train both high level policy and low level policy.

The high level policy is charged for choosing a feasible demonstrated observation depending on current observation so low level policy is capable of achieving that sub-goal in certain steps, i.e. five steps in our experiment. To represent the $o_g$ among $D$, we select a 2-dim action space for high level policy.
The high level policy's action consists of two rates between 0 and 1, the first dimension of action stands for which trajectory in $D$ is chosen, and second action dimension is the index of observation chosen in the trajectory.
In this way, the sub-goal can be formed as:
\begin{equation}
	\begin{array}{c}
		i = a_{h}^1 \times N   \\
		j = a_{h}^2 \times T_i \\
		o_g = d_j^i = \pi_{high}(a_{h}^1,a_{h}^2|o_t;\theta_h)
	\end{array}
\end{equation}
where the $N$ is the number of all demonstrated trajectories and $T_i$ means the length of trajectory $i$.

The low level policy is focused on interacting with the environment and find a way to achieve the sub-goal provided by the high level policy in certain steps. It takes $\{o_g,o_t\}$ as input and output $a_t$ as inter-action that interacts with the environment. The low level policy can be viewed as
\begin{equation}
	a_t = \pi_{low}(a_t|o_t,o_g;\theta_l)
\end{equation}
In this way, the low level policy and the high level policy together form the overall strategy. By flexibly selecting the observation in the demonstration as the sub-goal, the target task can be completed step-by-step.
The structure of our overall policy is shown in Fig. \ref{structure}. The training process is shown in Algorithm. \ref{algo}. 

\begin{algorithm}[h]
	\caption{HILONet}
	Initialize $\pi_{high}(a_{h}^1,a_{h}^2|o_t;\theta_h)$,$\pi_{low}(a_t|o_t,o_g;\theta_l)$.

	Collect $D$ using expert policy in the environment.

	Initialize replay buffer $R_{h}$, $R_{l}$.

	Initialize Q-Network $Q_{\theta_{h}}$ and target Q-Network $Q_{\theta_{h}'}$, Q-Network $Q_{\theta_{l}}$ and target Q-Network $Q_{\theta_{l}'}$.
	\begin{algorithmic}[1]

		\For{n steps = 1 to T}

		\State $\pi_{high}$ takes action $o_g = \pi_{high}(a_{h}^1,a_{h}^2|o_t;\theta_h)$.

		\State $\pi_{low}$ takes action $a_t = \pi_{low}(a_t|o_t,o_g;\theta_l)$.

		\State Get $<(o_t, o_g), a_t, r_l, o_{t+1}>$ and $<o_t, o_g, r_h, o_{t+1}>$.

		\If{$o_{t+1}$ in the $D$ and $o_{t+1}$ is not $o_g$}

		\State Replace $o_g$ by $o_{t+1}$ in both

		\State $<(o_t, o_g), a_t, r_l, o_{t+1}>$ and $<o_t, o_g, r_h, o_{t+1}>$.

		\EndIf

		\State Store $<(o_t, o_g), a_t, r_l, o_{t+1}>$ in $R_l$.

		\State Store $<o_t, o_g, r_h, o_{t+1}>$ in $R_h$.

		\State Sample a mini-batch $B_h$ and $B_l$ from $R_{h}$, $R_{l}$.

		\If{n mod TimeDelay}

		\State Perform a gradient decent step on $\pi_{high}$.

		\State Perform a gradient decent step on $Q_{\theta_{h}}$.

		\If{n mod TargetUpdate $\times$ TimeDelay}.

		\State Update $Q_{\theta_h'}: \theta_h' \gets \theta_h$.



		\EndIf
		\EndIf

		\State Perform a gradient decent step on $\pi_{low}$.

		\State Perform a gradient decent step on $Q_{\theta_{l}}$.



		\If{n mod TargetUpdate}

		\State Update $Q_{\theta_l'}: \theta_l' \gets \theta_l$.

		\EndIf

		\EndFor
		Start next episode.
	\end{algorithmic}
	\label{algo}
\end{algorithm}

\subsection{Reward Structure}
The rewards for both high level and low level policy are designed based on observation information as there is no action label available.
For low level policy, the insight is simple, that we want the reward to encourage the agent to achieve sub-goals. We can use the Euler distance of goal observation and current observation to do so directly. However, this reward can not reflect whether the sub-goal is achieved. For example, the reward increases similarly when agent approaching the sub-goal from a far distance or when agent is close to the target but missing the target slightly. To encourage agent achieving sub-goal, we add a sparse reward $r$ that is given only when the agent achieves the current sub-goal. We define if $|o_g-o_t|<\epsilon$ then we consider agent has achieved the sub-goal. The overall reward of low level policy can be viewed as:
\begin{equation}
	r_{low}(o_t, o_g) = \left\{
	\begin{array}{lr}
		-|o_g-o_t|^2     & \|o_g-o_t\|>\epsilon \\
		-|o_g-o_t|^2 + r & \|o_g-o_t\|<\epsilon \\
	\end{array} \right.
\end{equation}

As for the high level policy, we consider it should guide the low level policy to accomplish the specific tasks. In this way, the high level policy must do two tasks, finding a feasible sub-goal that agent can achieve and the sub-goal should gradually solve the specific tasks. The reward for the high level policy is designed based on these two proposals. First, we naturally use a sparse reward that given only when low level policy achieved the sub-goal. Then we add another reward related with which phase agent is right now.
This reward can be evaluated by $I(o_g) = I(d_j^i)$.
Because it represents the index of the reached expert state in the entire expert sequence. If the value of $I(o_g)$ is getting larger, means the agent is getting closer to the final goal. And since $I(o_g)$ is normalized between 0 and 1, it can be used among all trajectories even if they have the different number of steps to the end.
If we output sub-goal every $\Delta t$, we can use
\begin{equation}
	\Delta I(o_g) = I(o_g^{t}) - I(o_g^{t-\Delta t})
\end{equation}
as the reward to guide agent. 
Furthermore, we define $I(o_g) = 0$ if $o_{t}$ is not in expert trajectory, which can punish agent when it deviates from the expert trajectory. The overall reward of high level policy can be viewed as:

\begin{small}
	\begin{equation}
		r_{high}(o_g^{t}, o_g^{t-\Delta t}) = \left\{
		\begin{array}{lr}
			1+\alpha\cdot(I(o_g^{t}) - I(o_g^{t-\Delta t})) & \|o_g^i-o_t\|<\epsilon \\
			0                                      & \|o_g^i-o_t\|>\epsilon
		\end{array}\right.
	\end{equation}
\end{small}
where $\alpha$ is the multipliers to control the optimization ratio of these two parts of the reward.
And here we propose,

\textbf{Theorems 1:} \textit{By changing $\alpha$, we can control the behavior pattern of the learned policy, the higher $\alpha$ is, the learned policy would focus on following the trajectories as closely as possible. And the lower $\alpha$ is, the learned policy would focus on exploring more novel paths to the goal position.}


$Proof$ Firstly, we can have the optimal value at a state is given by the state-value function
\begin{equation}
	V^{*}\left(o_{t}\right)=\max _{\pi} \mathbb{E}\left[\sum_{t=0}^T \gamma^{t} r\left(o_{t}\right)\right].
	\label{eq2}
\end{equation}
where
\begin{equation*}
	r\left(o_{t}\right)=\max _{\pi} r\left(o_{t}, a_{t}\right).
\end{equation*}
we consider two optimum policies that can solve the tasks $\pi_1$ and $\pi_2$. Both policies are well-trained, which means the low level policies is the same, only high level policies are different. $\pi_1$ choose a path to reach the goal observation exactly as the expert does, while $\pi_2$ choose a total different path but reach the same goal position. The state-value function of these two policy are:
	\begin{equation}
		\begin{multlined}
			V_1^{*}\left(o_{t}\right)=\max _{\pi_1} \mathbb{E}\left[\sum_{t=0}^T \gamma^{t} r\left(o_{t}\right)\right]. \\
			V_2^{*}\left(o_{t}\right)=\max _{\pi_2} \mathbb{E}\left[\sum_{t=0}^T \gamma^{t} r\left(o_{t}\right)\right].
			\label{eq3}
		\end{multlined}
	\end{equation}
We substitute our reward $r_{high}$ into Eq. \ref{eq3} to obtain
	\begin{equation}
		\begin{multlined}
			V_1^{*}\left(o_{t}\right)=\mathbb{E}\left[\sum_{t=0}^T \gamma^{t} (1+ \alpha \cdot (I(o_t) - I(o_{t-\Delta t}))) \right] \\
			= \sum_{t=0}^T \gamma^{t} (1+\alpha \cdot \frac{\Delta t}{T})).
		\end{multlined}
		\label{eq4}
	\end{equation}
	\begin{equation}
		\begin{multlined}
			V_2^{*}\left(o_{t}\right)=\mathbb{E}\left[ \gamma^{T} (1+\alpha \cdot I(o_T)-0))\right] = (1+\alpha)\cdot\gamma^{T}.
		\end{multlined}
		\label{eq5}
	\end{equation}
We use Eq. \ref{eq4} minus Eq. \ref{eq5}, we have
	\begin{equation}
		\begin{multlined}
			\Delta V^{*}\left(o_{t}\right) = \sum_{t=0}^T \gamma^{t} (1+\alpha \cdot \frac{\Delta t}{T})) - (1+\alpha)\cdot\gamma^{T}
		\\ = \sum_{t=0}^T \gamma^{t} - \gamma^{T} + \alpha \cdot (\sum_{t=0}^T \gamma^{t} \cdot \frac{\Delta t}{T} - \gamma^{T})
		\\ = \sum_{t=0}^{T-1} \gamma^{t} + \alpha \cdot (\sum_{t=0}^T \gamma^{t} \cdot \frac{\Delta t}{T} - \gamma^{T}),
		\end{multlined}
		\label{eq6}
	\end{equation}
Then we have 
\begin{equation}
	\begin{multlined}
	 \sum_{t=0}^T \gamma^{t} \cdot \frac{\Delta t}{T} - \gamma^{T}
	 = \sum_{t=0}^{T} \gamma^{t} \cdot \frac{\Delta t}{T} + T \cdot \gamma^{T} \cdot \frac{1}{T}
	 \\ = \sum_{t=0}^{T} (\gamma^{t} \cdot \frac{\Delta t}{T} - \gamma^{T} \cdot \frac{1}{T}),
	\end{multlined}
	\label{eq7}
\end{equation}
As
\begin{equation}
	\begin{multlined}
	\gamma^{t}\geq \gamma^{T} ;
	\frac{\Delta t}{T} \geq \frac{1}{T},
	\end{multlined}
	\label{eq8}
\end{equation}
We have 
\begin{equation}
	\begin{multlined}
	 \sum_{t=0}^T \gamma^{t} \cdot \frac{\Delta t}{T} - \gamma^{T}
	 \geq 0.
	\end{multlined}
	\label{eq9}
\end{equation}
In this way, we prove that $\Delta V^{*}\left(o_{t}\right)$ is monotonically increasing with $\alpha$, when $\alpha$ is high, the return distinguish between $\pi_1$ and $\pi_2$ are large, which pushes the agent to follow the expert as close as possible. However, when the $\alpha$ is low, there is not much difference in the return of $\pi_1$ and $\pi_2$, which makes the agent tend to explore more novel path, as the novel path can increase the chance of obtaining rewards.


In summary, our hierarchical framework method has three advantages. First of all, comparing to another reward engineering imitation learning from observation methods, our method can plan dynamically thus it can solve the non-time-aligned problem. 
Furthermore, the reward structure of our method can adopt in all kinds of environments, no matter the tasks are described by a single goal observation or a sequence of key observations.
Additionally, our method can use information from multiple trajectories simultaneously, while most reward engineering methods can only imitate one trajectory if it imitates an expert step-by-step.

Secondly, our method is based on reward engineering, so comparing with adversarial methods, our method has access to observation information directly, which can offer more dense reward and will not miss any important information in demonstrated trajectories. 

The third advantage is that using a hierarchical framework policy can naturally divide a complex task into two more straightforward tasks, which will accelerate learning in sequential decision-making tasks.
\begin{figure}[t]
	\centering
	\includegraphics[width=1.0\linewidth]{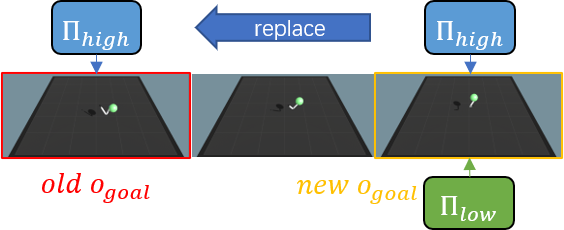}
	\centering
	\caption{Hindsight Transitions. Once low level policy achieves expert trajectories observations(but not the original sub-goal), we can use hindsight methods to replace sub-goal(high level policy action) by the new one.}
	\label{fig_hind}
\end{figure}
\subsection{Methods to Overcome the Non-stationarity in Hierarchical Framework}
\label{nonstationary}
In this section, we will introduce three methods that overcome the non-stationarity in the hierarchical framework so that we can increase sample efficiency.
\subsubsection{Hindsight Transitions}
Non-stationarity of hierarchical framework mainly comes from the low level policy. A transition collected when the low level policy is not well-trained can be useless. As the low level policy is changing that even giving the same sub-goal, the low level policy will execute a different transition. Moreover, this will change the distribution of reward for high level policy.

We can overcome this problem by using a hindsight replacement method, as shown in Fig. \ref{fig_hind}. Once the low level policy achieves an observation in expert trajectories, we consider the low level policy finishes the sub-goal offered by the high level policy. In this way, we can take this transition as a stationary transition generated by a well-trained low level policy. 


To achieve this, we need to replace the original transitions of high level policy to a hindsight one. If we have $\{o_t, o_g = \pi_{high}(o_t), r_{high}(o_{t+\Delta t},o_g), o_{t+\Delta t}\}$ as original transition. Once $o_{t+\Delta t}$ is in expert trajectories, we can use hindsight methods to replace sub-goal(high level policy action) by $o_{t+\Delta t}$, this means our high level policy's action chose $o_{t+\Delta t}$ as sub-goal at the first place, and our low level policy also use the best policy to achieve that sub-goal. So, the hindsight transition can be viewed as $\{o_t, o_g = o_{t+\Delta t}, r_{high}(o_{t+\Delta t},o_g), o_{t+\Delta t}\}$. By now, we have stationary hindsight transitions in high level policy replay buffer. However, not every transition can be replaced by stationary hindsight transition, for $o_{t+\Delta t}$ is not necessary within expert trajectories. We consider other methods to overcome this problem which will show below.

We also use hindsight transitions in low level policy replay buffer. The reason is that we could use failed trajectory as success one offering the agent more dense reward. Assuming high level policy outputs sub-goal every $\Delta t$. In this way, if we have $\{o_t, \pi_{low}(o_t, o_g), r_{low}(o_{t+1},o_g), o_{t+1}\}$ as original transition of low level policy. Once $o_{t+\Delta t}$ is in expert trajectories, we can use hindsight methods to replace sub-goal $o_g$ by $o_{t+\Delta t}$ like we did in high level policy. Then, the hindsight transition of low level policy is $\{o_t, \pi_{low}(o_t,o_{t+\Delta t}), r_{low}(o_{t+1},o_{t+\Delta t}), o_{t+1}\}$, it means our low level policy using a well-trained policy to achieve sub-goal that offered by high level policy instead of achieving a wrong observation.

To summarize, we use two different hindsight trajectory replacement methods in both high and low level policy.
These methods can overcome the non-stationarity of the hierarchical framework in high level policy.
Furthermore, we use the hindsight trajectory replacement method to increase the sample efficiency of the low level policy.
These methods can overcome the non-stationarity of the hierarchical framework and increase the sample efficiency of our method.

\subsubsection{Asynchronous Delayed Update}
\label{adu}
Since we know, the non-stationarity of hierarchical framework mainly comes from the imperfect low level policy. We can take measures to ensure that low level policy is trained firstly. As the better trained low level policy can significantly reduce non-stationary transitions in high level policy replay buffer.
To achieve this, we use a time-delayed training process to do so. The low level policy is trained every step while high level policy is trained with a delay, for instance, double steps.

The idea of time delay has shown its effectiveness in TD3\cite{dankwa2019twin}. TD3 uses it to train the actor and critic networks because it makes sense to train the actor network after the critic network is trained more accurately. In our case, this makes even more sense, because our high and low level policy are not facing similar tasks like actor and critic network in TD3. As the hierarchical structure divides the overall complex task into two simpler tasks, our high level policy does not need to interact with the environment. This makes its task not so hard to learn that even if we delay its training process, it can still learn a good enough policy.

\subsubsection{Smaller Experience Replay Buffer Size}

The last method to reduce the non-stationarity of the hierarchical framework is keeping our high level policy experience replay buffer size small. This can be easily explained that smaller size means using less old experience which we already know is bad for high level policy training. However, we just used a relatively small experience replay buffer, which does not have a significant impact on off-policy sample efficiency.
Furthermore, as we discussed in earlier section, the high level policy has an easier task to learn, which indicating using a smaller replay buffer does not do much harm to the learning process.

\begin{figure}[htbp]
	\centering
	\includegraphics[width=0.49\textwidth]{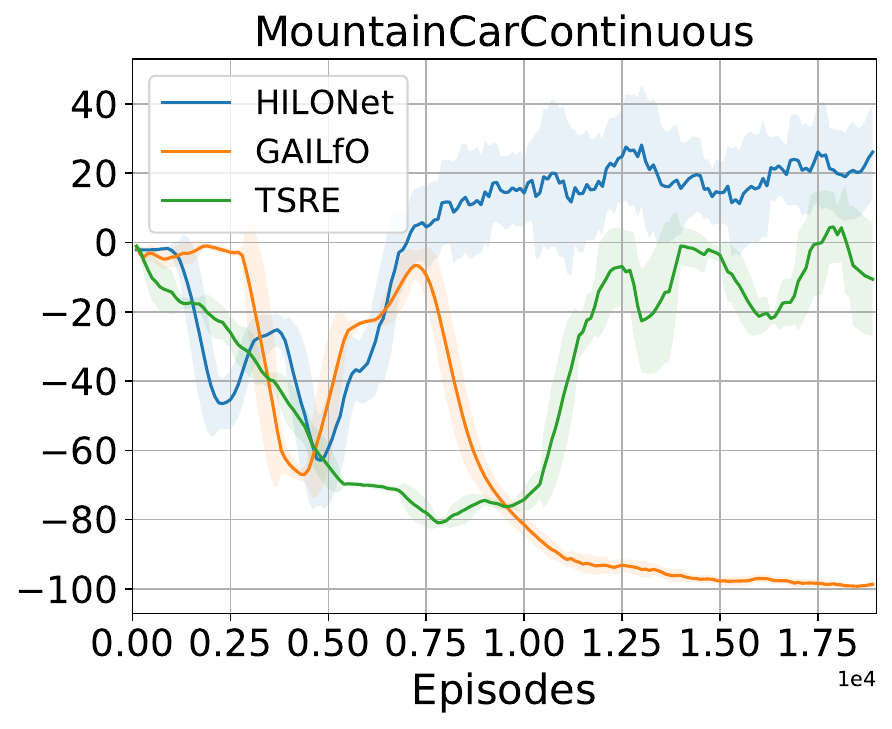}
	\caption{Performances in MountainCar environment.}
	\label{fig_mc}
\end{figure}
\begin{figure}[htbp]
	\centering
	\includegraphics[width=0.49\textwidth]{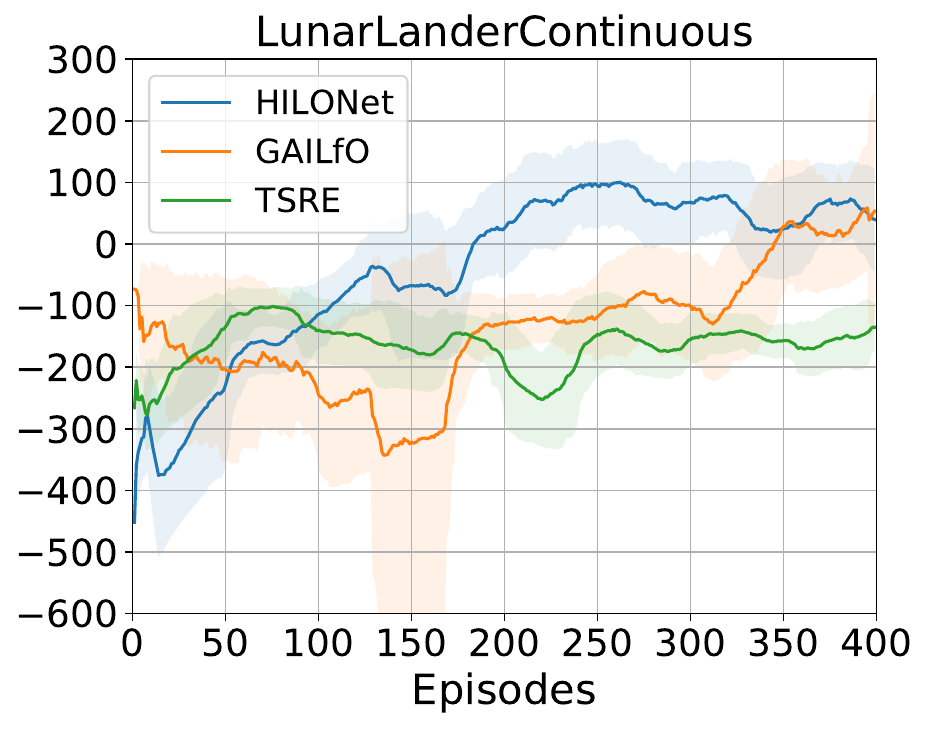}
	\caption{Performances in LunarLander environment.}
	\label{fig_ll}
\end{figure}

\begin{figure*}[htbp]
	\centering
	\includegraphics[width=0.85\textwidth]{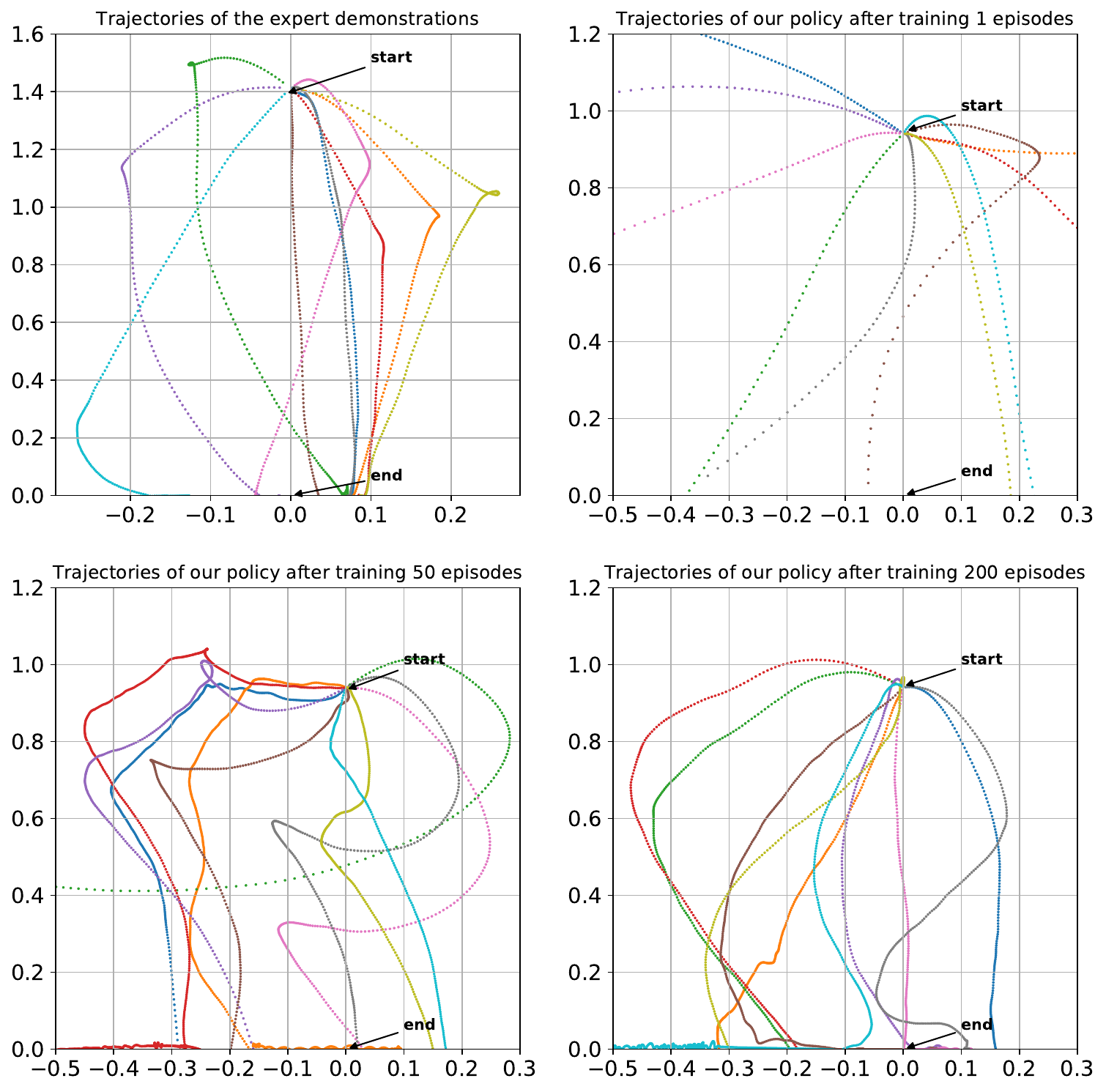}
	\caption{ Plots show the (x, y)-trajectories in LunarLander. All trajectories start at the start point and land near the endpoint, each plot contains ten trajectories, for instance. (a) shows trajectories of the expert demonstrations. (b)-(d) display the trajectories of our policy after training 1, 50 and 200 episodes. Over time, the high level policy learns to choose sub-goals that are not from a single trajectory. And low level policy learns to achieve the sub-goals.}
	\label{fig_adv}
\end{figure*}

\section{Experiments}
We evaluate our method in five different environments to show how our method benefits imitation learning in non-time-aligned environments. Especially, we are interested in answering two questions: (1) can our method choose the reachable observation from demonstration to solve these two kinds of non-time-aligned tasks? And (2) can it be compared with other state-of-the-art algorithms?

\subsection{Simulated Tasks}
We choose MountainCar, LunarLander, Swimmer in Gym\cite{brockman2016openai} and Reacher, 3Dball in ml-agents\cite{nandy2018unity}. All environments are shown in Fig. \ref{fig_env}. These five environments are all non-time-aligned and represent different characteristics, including two kinds of tasks can be effectively described by a single goal observation or a sequence of key observations rather than a single goal.
Specially, the MountainCar and LunarLander can be viewed as a single goal environment, while the Swimmer and 3Dball can be viewed as a sequence of goal environment. And Reacher is more complex that it can be viewed as a mixing environment. More details about environments will discuss below.
Experiments in these environments can show the universality of our method.

MountainCar: The goal is to have the car reach the target point. The target is on top of a hill on the right-hand side of the car.
If the car reaches it or goes beyond, the episode terminates. This means the number of steps of each episode is changing, which is a typical non-time-aligned environment. On the left-hand side, there is another hill. Climbing this hill can be used to gain potential energy and accelerate towards the target. On top of this second hill, the car cannot go further than a position equal to -1, as if there was a wall. Hitting this limit does not generate a penalty.

LunarLander: The goal is to land on a specific position without crashing.
The landing pad is always at coordinates (0,0). If lander moves away from the landing pad, it loses reward back. The episode finishes if the lander crashes or comes to reset, receiving additional -100 or +100 points. Landing outside the landing pad is possible. Fuel is infinite, so an agent can learn to fly and then land on its first attempt. These feature make sure that the agent can choose different path to the goal position, which is also a typical non-time-aligned environment.

Reacher: The agent must move their hand to a constantly moving goal location, and keep it there. The agent is a double-jointed arm which can move to target locations. The target location is a constantly moving location in 3D space, which makes the agent must follow its trajectory. 
This environment is different from the other environments. The moving target points bring an enormous challenge, as the first phrase of reaching the goal position can be viewed as a single-goal task and the second phrase of keeping moving within the goal area can be viewed as a sequenced-goal task.

3Dball: The agent must balance the ball on it's head for as long as possible.
This goal makes the environment different from the single-goal ones, as it not a reaching given goal point task which the goal point is usually the last frame of demonstration. This task needs to follow a special action pattern to keep a requested status, which is not directly related to the last frame of demonstration.

Swimmer: This task involves a 3-link swimming robot in a viscous fluid, where the goal is to make it swim forward as fast as possible, by actuating the two joints. Like 3Dball, this task requires the agent to arrive at the orderly goal frames in order to achieve high speed.

\subsection{Effect of Hierarchical Imitation Learning From Observation}

In this part, we evaluate the effect our Hierarchical Imitation Learning From Observation(HILONet) in these five environments. For comparison, we evaluate several state-of-the-art imitation learning algorithms, including adversarial methods like GAIL, which using a specific class of cost functions. This allows for the use of generative adversarial networks in order to do apprenticeship learning. And GAILfO\cite{torabi2018generative}, which is similar to GAIL but only using observations. 
Meanwhile, we evaluate a baseline reward engineering method, which imitates the expert observation step-by-step. This method learns to reach the observation of the same time step of the demonstrated trajectory at each time step.
We call it time-sequential reward engineering(TSRE). Notable, our method, GAILfO and reward engineering baseline are using only expert observations, but GAIL have access to the demonstrator's actions.

We implement HILONet using DDPG algorithm. In the implementation of DDPG, the policies are all parameterized by a two-layer full connected network with 64 units per layer and ``relu'' active function and are initialized randomly. The high level policy acts every five steps. The $\alpha$ is 0.5 in single-goal task and 10 in the sequenced-goal task. We use precisely the same parameters for all compared algorithms. As for demonstrations, we use 20 trajectories in MountainCar, 30 trajectories in LunarLander, Swimmer, 3Dball and Reacher. However, TSRE can only use one of these trajectories as it can only follow one expert step-by-step. All demonstrations are collected by pre-trained expert policies which are trained with extinct reward.
All tests are evaluated over three seeds and using 20 or 30 experts' trajectories in all environments. For all experiments, we utilize the environmental return as the performance of each method, which is the y-axis in all plots.

\begin{table*}
	\centering
	\begin{tabular}{ccccccc} \hline
		   &                  & MountainCar           & LunarLander            & Reacher              & 3Dball               & Swimmer                \\ \hline
		IL & GAIL             & 56.3$\pm$11.5         & 149.9$\pm$38.5         & 1.32$\pm$0.49        & 3.1$\pm$0.3          & 43.3$\pm$1.7           \\
		\hline
		ILfO
		   & GAILfO           & -98$\pm$1.5           & 32.0$\pm$73            & 0.2$\pm$0.1          & 0.18$\pm$0.02        & 25.6$\pm$2.4           \\
		   & TSRE             & -3.3$\pm$11.2         & -148.9$\pm$99          & 0.73$\pm$0.3         & 0.3$\pm$0.0          & 17.3$\pm$2.2           \\
		   & \textbf{HILONet} & \textbf{22.9$\pm$9.7} & \textbf{61.0$\pm$37.7} & \textbf{1.1$\pm$0.5} & \textbf{6.5$\pm$3.4} & \textbf{38.5$\pm$1.95} \\
		\hline
	\end{tabular}
	\caption{The final performance of all four methods mentioned before. Notably, the GAIL is an imitation learning method as it has access to action labels, while others is learning from observation methods that utilize pure observation without action label from demonstrations.}
	\label{table}
\end{table*}


\subsubsection{Tasks with Single Goal}
Fig. \ref{fig_mc} and Fig. \ref{fig_ll} depicts the performance of the agents trained with all methods in MountainCar and LunarLander. In all cases, our method outperforms the GAILfO and TSRE. These two algorithms are our main comparison algorithms because they are the same as ours belonging to ILfO algorithms. We find GAILfO has a much poorer performance in MountainCar, which learned a sub-optimal policy that only focuses on going back. This is because the backward acceleration samples account for the majority in the demonstrations, we think that in the process of GAILfO learning, it pays too much attention to these samples and ignores the small part of the forward acceleration samples(including final goal position) that are important to complete the task. However, unlike GAILfO, the reward of HILONet makes the agent to reach the desired goal observations directly. Therefore, there will not be a situation where a small part of important samples is ignored like GAILfO. We also notice that the TSRE can not learn a decent policy in both two environments, as it can only learn from a single trajectory which limits the possibility of finding a novel way to the goal position. On the contrary, HILONet can dynamically choose sub-goals at each time step to explore more paths. As a result, HILONet has the best performance among all methods.

\begin{figure}[htbp]
	\centering
	\includegraphics[width=0.49\textwidth]{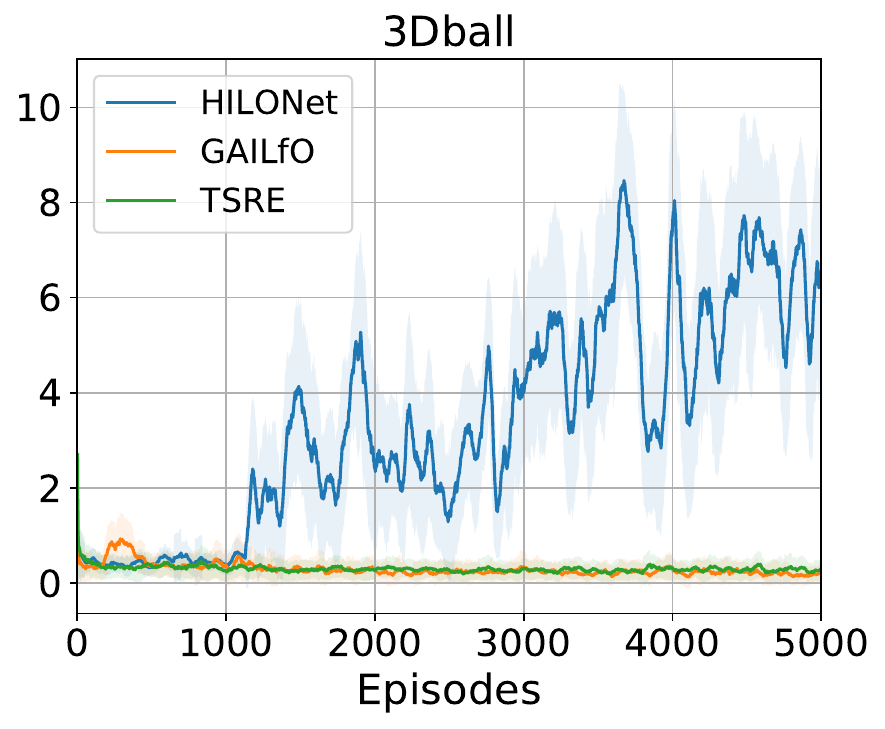}
	\caption{Performances in 3Dball environment.}
	\label{fig_3d}
\end{figure}
\begin{figure}[htbp]
	\centering
	\includegraphics[width=0.49\textwidth]{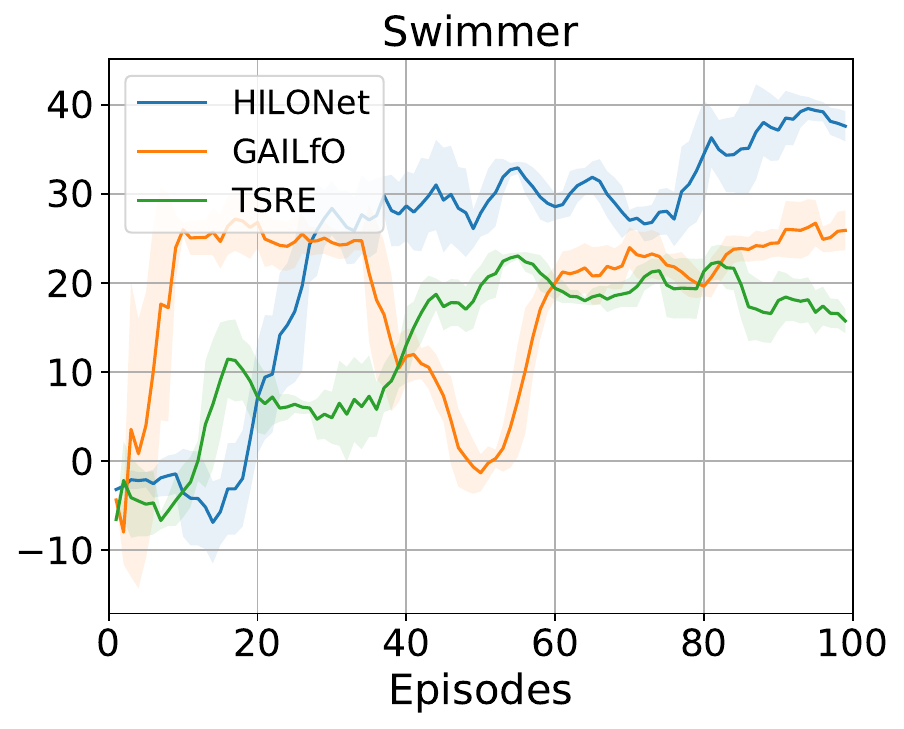}
	\caption{Performances in Swimmer environment.}
	\label{fig_sw}
\end{figure}

\subsubsection{Visualization of Learned Trajectory}
In order to show the novel paths learned to solve single-goal tasks, we plot expert trajectory and our policy's trajectory in LunarLander for comparison. Each image contains ten trajectories, for instance. We show our policy's trajectories from beginning to the end of the training so that we can observe the learning process. The result is shown in Fig. \ref{fig_adv}. We find that our policy's trajectory is not exactly the same as the expert's. This means our method learns some new paths to the final goal.

\subsubsection{Tasks with Sequenced Goals}
Fig. \ref{fig_3d} and Fig. \ref{fig_sw} depicts the performance of the agents trained with all methods in 3Dball and Swimmer.
The results show that our method learns the expert action pattern from the order of demonstration frames.
In 3Dball and Swimmer environments, tasks can not be solved by purely achieve the final observation of a demonstrated trajectory. Agents must go through a series of specific observations to solve the task. And our method can learn to find these specific observations by following the order of demonstration frames, which is exactly our high level policy's task when $\alpha$ is large. This conclusion is in the line with our theoretical proof. Moreover, we notice that TSRE has a comparable performance to GAILfO in sequenced-goals tasks. We believe that the reason is the sequenced-goals tasks requiring to follow the demonstrated trajectory strictly, which makes TSRE more advantageous than GAILfO, even if it only utilizes one trajectory.
\subsubsection{Mixing Tasks}
\begin{figure}[t]
	\centering
	\includegraphics[width=0.49\textwidth]{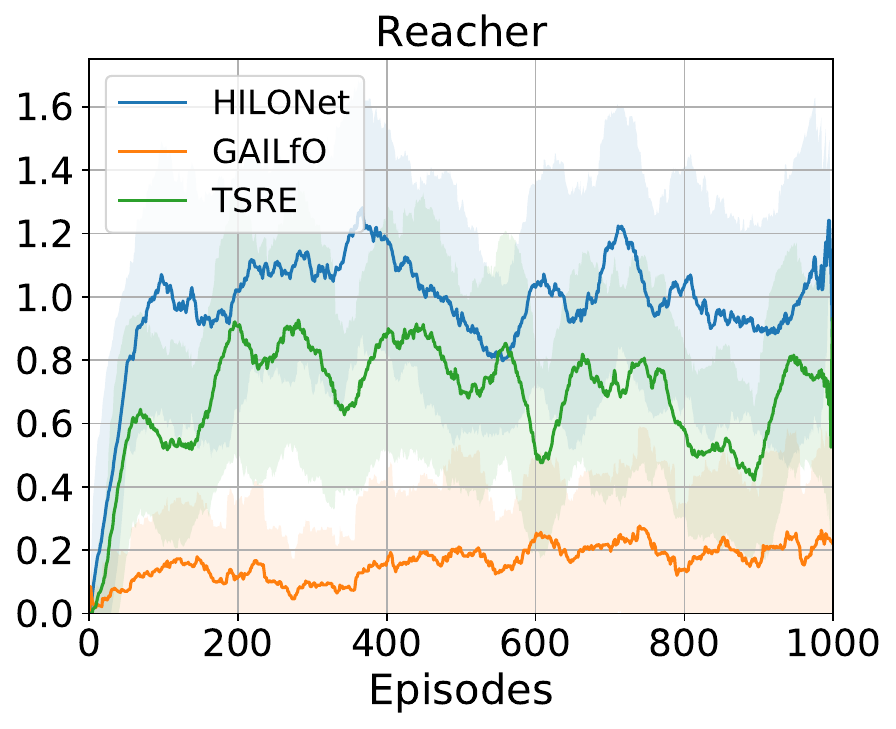}
	\caption{Performances in Reacher environment.}
	\label{fig_re}
\end{figure}

\begin{figure}[t]
	\centering
	\includegraphics[width=0.49\textwidth]{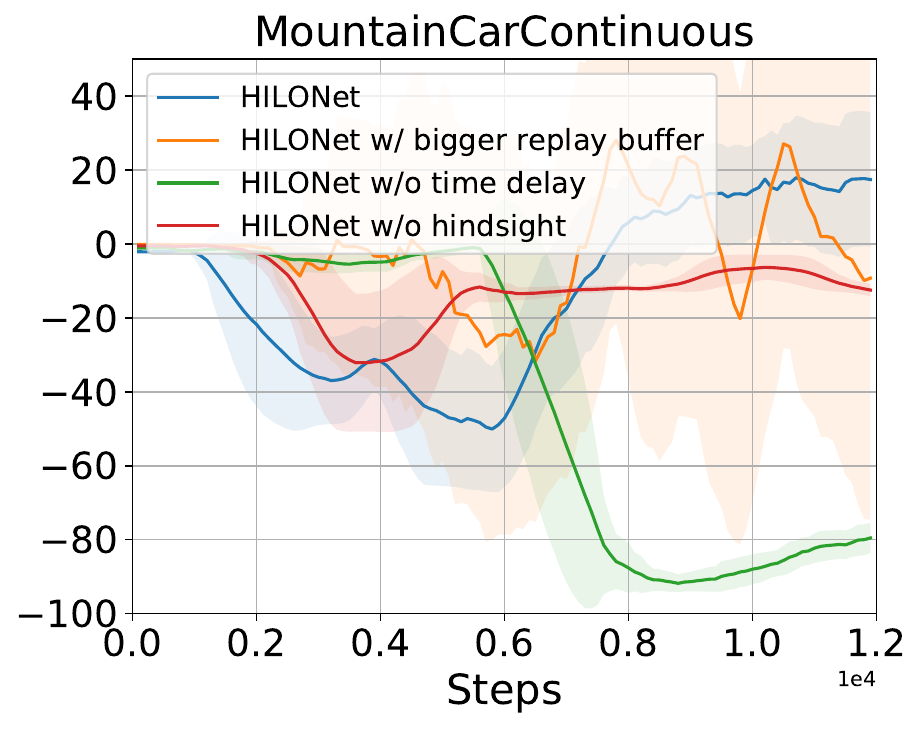}
	\caption{Ablation in MountainCar environment.}
	\label{fig_amc}
\end{figure}
\begin{figure}[t]
	\centering
	\includegraphics[width=0.49\textwidth]{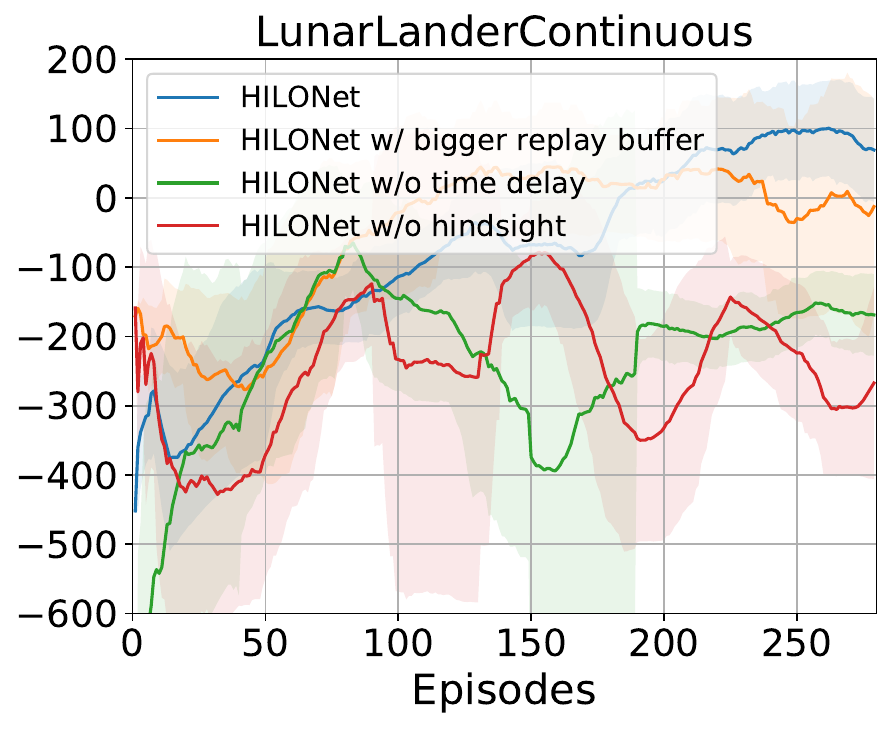}
	\caption{Ablation in LunarLander environment.}
	\label{fig_alc}
\end{figure}
\begin{figure}[t]
	\centering
	\includegraphics[width=0.49\textwidth]{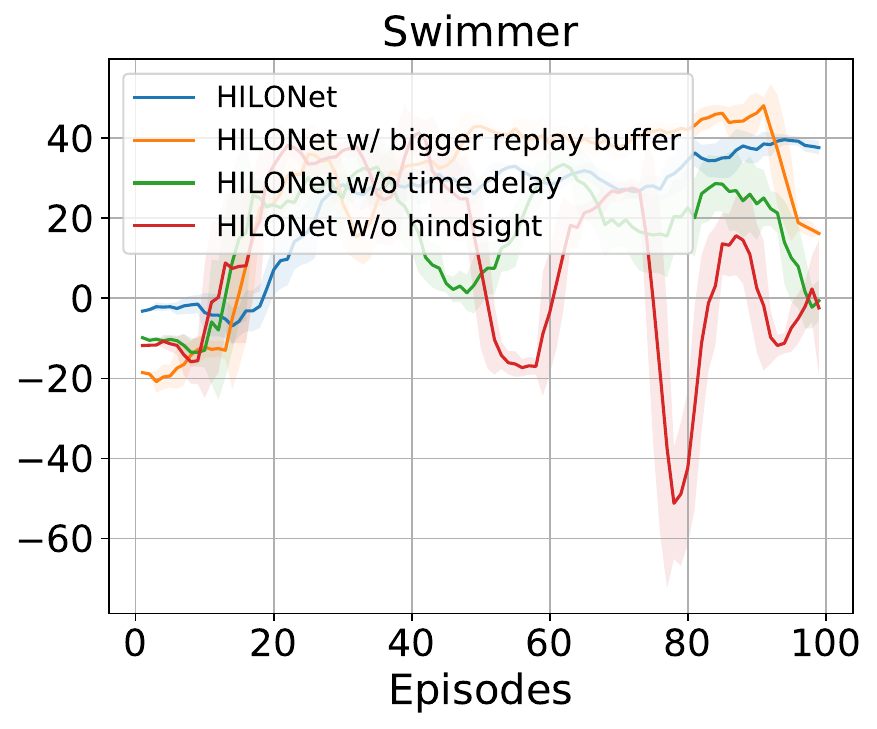}
	\caption{Ablation in Swimmer environment.}
	\label{fig_asw}
\end{figure}
Fig. \ref{fig_re} depicts the performance of the agents trained with all methods in Reacher. As Reacher is a mixing task, the result is similar to the former ones. We find our method still outperforms all compared methods when the environment involves both two kinds of tasks.

Finally, we show the final performance of all four methods in Table. \ref{table}. As for GAIL, it performs best because it has access to action labels from the expert. However, in all these five environments, our method is not far worse than GAIL, HILONet even gets a higher score in environment 3Dball.

In summary, all experiment results show that our method has a better performance compared with either TSRE or adversarial method GAILfO. As a unique hierarchical structure is used, we can prove that the hierarchical structure does have advantages when performing imitation learning in non-time-aligned environments by achieving dynamic planning.

In addition, the hierarchical structure can solve the non-time-aligned problem in both two kinds of environments, including single-goal task and sequenced-goals task. Our method can learn new paths to the goal in tasks described by a single goal observation and follow the demonstrated trajectories as close as possible to learn the special action pattern in sequenced-goals tasks. 
As a result, HILONet has strong versatility.

\subsection{Ablation Study}
In this setting, we test how different ways of overcoming non-stationarity affect the agent's performance. Firstly, we test our method training without hindsight transition replacement in both high level policy and low level policy. Then, we find out how the time delay method affects the training process. At last, we test our method with a bigger replay buffer size, e.g. twice as big as the original one.

These results in Fig. \ref{fig_amc}, Fig. \ref{fig_alc} and Fig. \ref{fig_asw} shows that all these three ways that we proposed to overcome non-stationarity in the hierarchical framework works. Especially, methods without hindsight transition replacement or time delay training can hardly even get out of ground level. As for using a bigger replay buffer size, the influence is not as huge as others. However, we can still observe a significant decay.
The reason for this is that hindsight transition replacement or time delay training changes the distribution of the sample, so it has a greater impact on the learned policy. In contrast, changing the size of replay buffer only changes the learning hyperparameters, and the impact will not be so large. 

\section{Conclusion}

In this paper, we introduced a new imitation learning from observation method, hierarchical imitation learning from observation (HILONet), using hierarchical reinforcement structure to choose observations from expert trajectories' observations as goals. By achieving these goals, our method can imitate expert with only observations offered.
Furthermore, we propose a reward structure that can control the behavior pattern of learning policy whether to explore more or mimic more. 
In this way, our method has the ability to solve tasks with single goal position and tasks described by a sequence of key observations. 
Additionally, we propose three different ways to overcome the non-stationarity problem in hierarchical structure to increase sample efficiency. We evaluate the method with extensive experiments based on five different environments, including both these have a single goal position and those do not have a specific target. The result shows that HILONet can solve all kinds of tasks and improves the training procedure of imitation learning. It outperforms GAILfO and reward engineering baseline and achieves nearly the same performance as GAIL in all environments. Furthermore, we test the effect of these three ways of overcoming non-stationarity. As a result, we find they improve the effectiveness of our method.

For further research, we will attempt to use a multi-level hierarchical structure, which is believed can divide complex tasks into more simple ones. This may be the key to a general imitation learning algorithm that can be used in robot control, autonomous driving and other fields.

\bibliographystyle{IEEEtran}
\bibliography{hilfo.bib}



\end{document}